\documentclass[10pt,twocolumn,letterpaper]{article}

\usepackage[pagenumbers]{cvpr} 
\usepackage[accsupp]{axessibility}  

\definecolor{cvprblue}{rgb}{0.21,0.49,0.74}
\usepackage[pagebackref,breaklinks,colorlinks,allcolors=cvprblue]{hyperref}

\usepackage{multirow}
\usepackage{makecell}
\usepackage{pifont}
\usepackage{stmaryrd}
\usepackage{colortbl}

\title{FIRE-CIR: Fine-grained Reasoning for Composed Fashion Image Retrieval}

\author{François Gardères$^{1, 2}$, Camille-Sovanneary Gauthier$^{1}$, Jean Ponce$^{2, 3}$, Shizhe Chen$^{2}$\\
{\normalsize $^1$ Louis Vuitton}\\
{\normalsize $^2$ Inria, École normale supérieure, CNRS, PSL Research University}\\
{\normalsize $^3$ Courant Institute of Mathematical Sciences and Center for Data Science, New York University}\\
{\tt\small firecir.contact@gmail.com}
}

\newcommand{\cmark}{\ding{51}}
\newcommand{\xmark}{\ding{55}}

\begin{document}
\maketitle
\begin{abstract}

Composed image retrieval (CIR) aims to retrieve a target image that depicts a reference image modified by a textual description. While recent vision-language models (VLMs) achieve promising CIR performance by embedding images and text into a shared space for retrieval, they often fail to reason about what to preserve and what to change.
This limitation hinders interpretability and yields suboptimal results, particularly in fine-grained domains like fashion.
In this paper, we introduce FIRE-CIR, a model that brings compositional reasoning and interpretability to fashion CIR. Instead of relying solely on embedding similarity, FIRE-CIR performs question-driven visual reasoning: it automatically generates attribute-focused visual questions derived from the modification text, and verifies the corresponding visual evidence in both reference and candidate images. To train such a reasoning system, we automatically construct a large-scale fashion-specific visual question answering dataset, containing questions requiring either single- or dual-image analysis.
During retrieval, our model leverages this explicit reasoning to re-rank candidate results, filtering out images inconsistent with the intended modifications. Experimental results on the Fashion IQ benchmark show that FIRE-CIR outperforms state-of-the-art methods in retrieval accuracy. It also provides interpretable, attribute-level insights into retrieval decisions.
Code:
\href{https://fgxaos.github.io/firecir-paper-website}{https://fgxaos.github.io/firecir-paper-website}.

\end{abstract}    
\section{Introduction}
\label{sec:intro}

\begin{figure}[t]
  \centering
    \includegraphics[width=\linewidth]{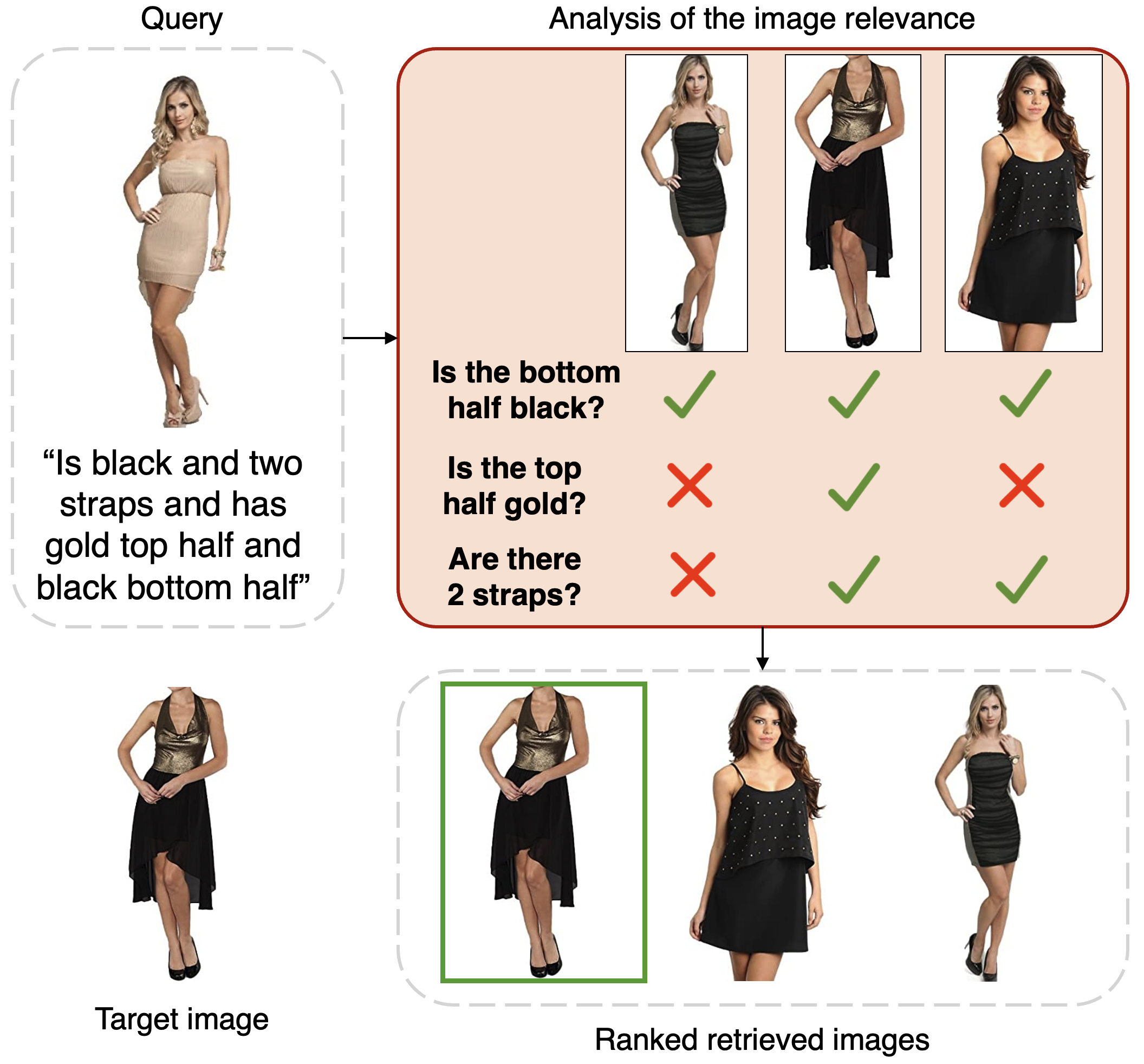}

   \caption{Given a CIR query, our model analyzes the presence of the requested changes in a list of candidate images, to re-rank them depending on their relevance with respect to the modification text. The target image is framed in green.}
   \label{fig:teaser_figure}
\end{figure}

The rise of multi-modal search has transformed user interaction with visual content, allowing for complex queries that combine images and text. 
This is particularly impactful in fashion e-commerce, where users leverage visual and textual cues to precisely refine product searches. Among these interactions, composed image retrieval (CIR) 
has emerged as a key task: given a reference image and a modification text (\eg, ``this shirt, but with longer sleeves"), the goal is to retrieve the image that best reflects the intended change.

Current CIR methods~\cite{baldrati2022conditioned, liu2024bi, bai2024sentence, wen2023target, feng2024improving, garderes2025facap, yang2025detailfusion} often build on vision-language models (VLMs) like CLIP~\cite{radford2021clip} and project both the text and the images into a shared embedding space for multimodal retrieval.
While effective, these approaches face two critical limitations, especially when applied to the fine-grained fashion domain. 
(i) Incomplete coverage: the methods struggle to accurately detect and fully leverage all intricate details contained within the textual and visual input data for result retrieval. 
(ii) Suboptimal fusion: the fused multimodal query often retains information from the reference image that is incompatible with the constraints specified in the modification text, consequently elevating the rank of results that are only partially relevant.

To mitigate these issues, a line of work~\cite{miech2021thinking, liu2024candidate} performs fine-grained matching using more detailed embedding representations. These methods are more precise but less efficient, so they are primarily used for re-ranking: a fast method outputs an initial ranking, and the top-retrieved results are then refined using the slower fine-grained matching method. 
Despite improving retrieval performance on fine-grained details, these models do not necessarily address the issue of results being partially incompatible with the modification text. 
In response, zero-shot CIR models~\cite{wu2025square, xiao2025setr, luo2025imagescope, lin2025mmembed, sun2024grb} leverage large pre-trained VLMs and their reasoning ability to address both issues. In spite of encouraging results in the general domain, they suffer from lower CIR performance in fashion, as their large pre-trained model may struggle with the specificities of fashion visual elements and vocabulary.
Thus, recent models~\cite{liu2025lamra, feng2025vqa4cir} fine-tune large VLMs on Fashion IQ~\cite{wu2020fashioniq}, a CIR fashion dataset to improve their knowledge on the target domain. However, due to a lack of detailed annotations, they rely on the hypothesis that the target image is the only relevant retrieved result. It is often not the case, as several candidate images can also be relevant to the given query. For example, querying a black T-shirt with longer sleeves could correspond to many different long-sleeved black T-shirts. Hence, these approaches suffer from approximate fine-tuning data, and lack fine-grained annotations to adapt to the fashion domain.

In this paper, we introduce FIRE-CIR, FIne-grained REasoning for Composed fashion Image Retrieval, a novel question-driven visual reasoning framework to address these core challenges. 
Instead of relying solely on embedding similarity, FIRE-CIR leverages explicit compositional reasoning by automatically decomposing the modification text into a set of attribute-focused visual questions, as illustrated in \Cref{fig:teaser_figure}.
These questions can either refer only to the candidate image, or require a comparison between the candidate and the reference images. 
To achieve superior performance in the fashion domain, we automatically construct a large-scale, fashion-specific visual question answering (VQA) dataset that incorporates both single- and dual-image analysis. This specialized dataset enables the fine-tuning of a VQA model to overcome the limitations of existing CIR data and precisely identify relevant images based on fine-grained visual elements.
By training a dedicated fashion VQA model to answer these queries, FIRE-CIR generates an interpretable reasoning trace that accurately measures a candidate image's compatibility with the query.
FIRE-CIR is designed as a plug-in reasoning method that can be seamlessly integrated with existing CIR models to perform re-ranking, effectively promoting candidates that are fully compatible with the textual modification. We demonstrate the efficacy of our framework on the Fashion IQ benchmark, showing that FIRE-CIR significantly improves retrieval accuracy and provides interpretable attribute-level insights into its decisions.

\noindent In summary, our key contributions are:
\begin{itemize}
    \item a new question-driven visual reasoning approach, which improves the relevance analysis of candidate images by enabling fine-grained comparison between the reference image and them;

    \item a fine-tuning approach based on a new automatically-annotated VQA dataset, to adapt reasoning to the target fashion domain even without detailed CIR annotations, while mitigating the impact of false negatives;
    
    \item a new method which can be plugged into any CIR model to improve their retrieval performance and obtain state-of-the-art results, by re-ranking retrieved images depending on their compatibility with the given modification text.
    
\end{itemize}
\section{Related Work}
\label{sec:related_work}

\subsection{Composed Image Retrieval}
\label{subsec:composed_image_retrieval}

Most CIR approaches \cite{baldrati2022conditioned, bai2024sentence, liu2024bi, zhao2024unifashion, yang2025detailfusion} learn a joint representation of the reference image and the modification text, which is then compared to the representation of the target images. These representations rely on large multi-modal embeddings, computed with models like CLIP~\cite{radford2021clip}, BLIP~\cite{li2022blip}, and BLIP-2~\cite{li2023blip2}. In order to specialize the embeddings and incorporate more information relevant to the retrieval task, they are fine-tuned on the target data domain. Simple fusion methods like embedding interpolation in CLIP4CIR~\cite{baldrati2022conditioned} achieve good retrieval performance on global concepts, but have issues with more precise textual or visual details. 
Indeed, despite having some detailed requirements in their modification texts, CIR datasets generally propose coarse annotations and lack precise details which could help guide the models' representation training. 

Recent approaches \cite{wen2023target, garderes2025facap, yang2025detailfusion} improve the ability of CIR methods to manipulate both coarse and precise elements by leveraging the implicit knowledge of large VLMs and by using complex representations which are able to incorporate more information at different scales. However, fusing the reference image and modification text representations tends to retain information from the image. A part of this visual information does not apply the changes requested in the text and may be incompatible with it. Thus, some undesired characteristics of the reference image are kept in the retrieved results, leading to highly-ranked images which are only partially compatible with the given modification text. Moreover, the feature-based comparison acts like a black box, as it does not explicitly specify which elements of the candidate image justify its retrieval rank.

\begin{figure*}[t]
  \centering
    \includegraphics[width=0.9\linewidth]{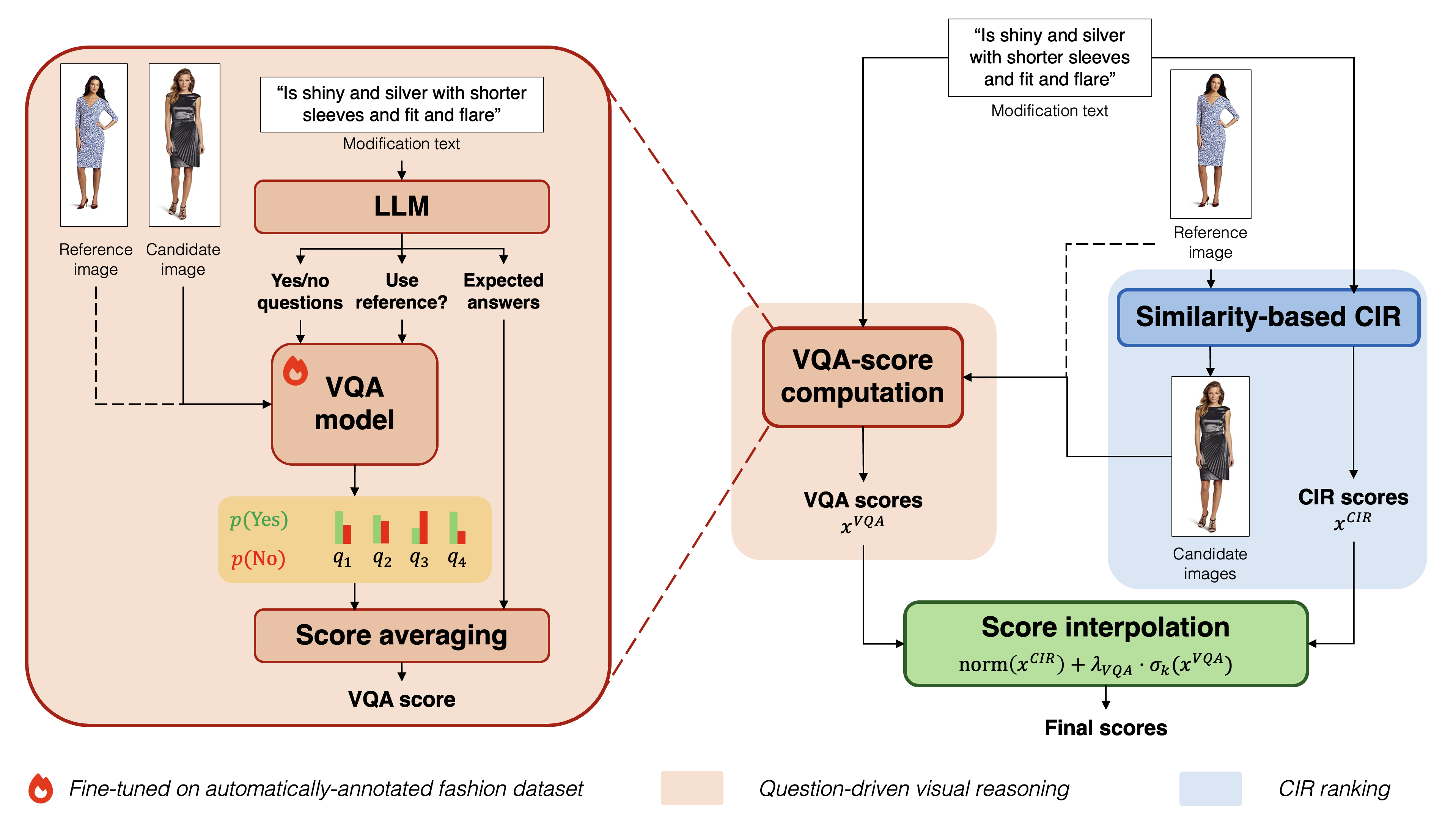}

   \caption{Overview of the FIRE-CIR model. Left: VQA score computation. The modification text is decomposed into a set of visual questions about the candidate image, which are answered using a VQA model. The predicted answers are then used to compute the VQA score, measuring the relevance of the candidate image with respect to the text. The set of questions and their answers give an interpretable insight in the relevance evaluation of each image. Right: CIR inference. Given the CIR query and a set of candidate images, FIRE-CIR computes the VQA score of each candidate image, and combines it with the score returned by another CIR method to refine the ranking of the retrieved images.}
   \label{fig:complete_pipeline_description}
\end{figure*}

\subsection{Fine-Grained Alignment for CIR}
\label{subsec:reranking_approaches}

Recent CIR methods have focused on two challenges: accurately matching fine-grained textual and visual concepts, and filtering out retrieved results which are incompatible with the modification text.

ReRanking~\cite{liu2024bi} and LamRA~\cite{liu2025lamra} train additional modules plugged into their VLM to better take into account fine-grained details in the text and images. Despite improving CIR performance, these models tend to still retain visual information incompatible with the modification text.

To address both challenges, a second category of methods directly uses a large multi-modal model to evaluate the relevance of a candidate image relative to a given modification text, benefitting from the model's reasoning capacity and its ability to analyze fine-grained details. Some methods \cite{xiao2025setr, lin2025mmembed, wu2025square} use a general prompt to evaluate if the analyzed candidate image is compatible with the modification text, either image-by-image like MM-Embed~\cite{lin2025mmembed} and SETR~\cite{xiao2025setr}, or by directly ranking a set of images like SQUARE~\cite{wu2025square}. 
However, by considering a single prompt, the model may focus on only some elements of the modification text and discard some precise details. This ranking process also lacks interpretability.

To mitigate this issue, GRB~\cite{sun2024grb}, ImageScope~\cite{luo2025imagescope} and VQA4CIR~\cite{feng2025vqa4cir} use the modification text to extract the different characteristics that the target image should possess. By decomposing the required changes into a list of specific constraints, a VLM can be used as a VQA model to evaluate the relevance of each candidate image in an interpretable manner. 
In the case of VQA4CIR~\cite{feng2024improving}, the VLM is adapted to the target domain with fine-tuning, improving its CIR performance. 
However, this fine-tuning process relies on the assumption that only the target image verifies all the constraints mentioned in the modification text. This is not necessarily the case, as there can be other candidate images fitting the given query: we call them ``false negatives".
Thus, the VQA model is implicitly fine-tuned on noisy data, which limits its ability to properly answer questions on fine-grained details and impacts its relevance scores.

Using an automatically-annotated VQA dataset makes our approach more robust to false negatives. 
It also enables single- and dual-image analysis, extending the range of constraints from the modification text that can be checked.

\subsection{Fashion Image Understanding}
\label{subsec:fashion_image_understanding}

The fashion domain has garnered a lot of interest for multi-modal tasks. Adapting models to this domain is challenging, due to its specific vocabulary and visual elements, as well as the importance of fine-grained details to distinguish similar products. Furthermore, this adaptation relies on multi-modal web-crawled datasets which often provide vague annotations describing only coarse elements. FashionVQA~\cite{wang2023fashionvqa} fine-tunes a VLM on a large-scale fashion dataset to improve its capacity to auto-label products, and uses the multi-modal embeddings for other downstream tasks. Similarly, several CIR works~\cite{han2022fashionvil, zhao2022progressive} have focused on improving the pre-training of large multi-modal models on fashion images, while FAME-ViL~\cite{han2023fame} and UniFashion~\cite{zhao2024unifashion} use multi-task learning to improve fashion image understanding.

Our approach also fine-tunes a VLM on a large-scale automatically-annotated dataset, but we incorporate questions on both images and image pairs, in order to train the model to reason on the given images. Additionally, we mitigate question formulation bias so that the model properly learns multi-modal interactions between text and image.
\section{Proposed Method}
\label{sec:method}

In this section, we introduce our model FIRE-CIR, which performs question-driven fine-grained visual reasoning to filter out retrieved results incompatible with the modification text, in an interpretable manner.
FIRE-CIR first converts the modification text into a set of attribute-focused visual questions, detailing the changes to apply. The correct application of the changes is then checked with a VQA model, which computes a text-relevance score. This score is finally used to refine the rank of the top-retrieved candidate images. An overview of the complete re-ranking process is illustrated in \Cref{fig:complete_pipeline_description}.

After formulating the CIR problem in \Cref{subsec:problem_formulation}, we present the visual question generation process in \Cref{subsec:criteria_extraction}. Our VQA model fine-tuning approach is then detailed in \Cref{subsec:vqa_finetuning}, enabling the evaluation of the relevance of a candidate image in a robust manner.
Finally, \Cref{subsec:score_merging} presents how FIRE-CIR can be leveraged for CIR inference. 

\subsection{Problem Formulation}
\label{subsec:problem_formulation}

Given a reference image $I_r$ and a modification text $T$, CIR aims to retrieve the correct target image $I_t$ among a set of candidate images $\mathcal{D} = \{I_0, ..., I_N \}$. This retrieval process computes a score $x_i$ for each candidate image $I_i$, which is used to rank them. Candidate images depicting the modifications mentioned in $T$ and the visual features of $I_r$ should be ranked among the top results, with $I_t$ having the highest score.

\subsection{Visual Question Generation}
\label{subsec:criteria_extraction}

The first step of FIRE-CIR converts the modification text into a list of questions about the changes to apply to the reference image, to check their correct application in each candidate image.
Contrary to VQA4CIR~\cite{feng2025vqa4cir}, we extract three pieces of information for each multi-modal query: (i) a yes/no question about a specific element in the given image, (ii) the answer compatible with the modification text, and (iii) whether the reference image is needed to check the change on the candidate image.

The yes/no questions are designed to cover all the visual elements corresponding to the changes mentioned in the modification text, with each question focusing on a single different concept. 

Additionally, we do not expect the answer to the question to always be ``Yes", especially when checking if an element has been removed from the reference image. The expected answer is used later in the inference process to check if the candidate image is compatible with the modification text.

Contrary to previous approaches, we detect whether the VQA question compares the candidate image to the reference image, for example when checking if an item is longer in the candidate image.
This process has two benefits: (i) it enables the VQA model to analyze and compare both reference and candidate images when needed, and (ii) when the question can be answered with just the candidate image, not only does it prevent information leakage from the reference image which could pollute the VQA prediction, but it also lets the VQA model process only the candidate image, hence improving the inference speed.

As this task is purely textual, we use an LLM to perform the question generation, and add an in-context example to guide the format of its outputs.

\subsection{Adapting the VQA model to Fashion Domain}
\label{subsec:vqa_finetuning}

Once the set of yes/no questions has been generated, we need to be able to predict their answer for each candidate image. Therefore, we need a VQA model able to not only precisely answer fashion-related questions, but also to analyze the reference and candidate images before comparing them. As these two tasks remain challenging for existing models, we fine-tune a large VLM to train this reasoning process and use it as a VQA model.

Fine-tuning requires a large set of questions referring either to one or two images, so we construct a new automatically-annotated VQA dataset.
We could use only the annotated target images, as their expected answer is already known thanks to our question-generating process in \Cref{subsec:criteria_extraction}. However, this would lead to a dataset with approximately $94\%$ ``Yes" answers. As explained in \cite{zhu2021vqapriors}, such a dataset would have a heavy language bias that VQA models could exploit to predict the answer only with the formulation of the question, hence bypassing the image analysis. To prevent this, for any question in our dataset, we sample different images so that the ``Yes" and ``No" answers are balanced.
As we do not have any information about the answer to the VQA questions for non-target images, we use a larger VLM to automatically annotate them. This enables us to improve the accuracy of our VQA model, without the need to use a larger and slower model at inference.

Thanks to this process, we obtain a VQA dataset with 413,848 question-image pairs, perfectly balanced in terms of ``Yes" and ``No" answers, and with about $32\%$ questions requiring a comparison between two images.
To retain the knowledge inherent to our large VLM while specializing it to the fashion domain, we use low-rank adaptation (LoRA)~\cite{hu2022lora} and fine-tune the model on our VQA dataset.

\subsection{CIR Inference}
\label{subsec:score_merging}
Given a CIR query and a candidate image, we first extract from the modification text a set $\mathcal{Q}$ of VQA questions. Then, for each element of $\mathcal{Q}$, we use the VQA model to compute the probability of the answer compatible with the modification text. We then average the probabilities over all the questions to obtain the VQA score of the candidate image, measuring its compatibility with the text constraints:
\begin{equation}
  x^{\text{VQA}} = \frac{1}{|\mathcal{Q}|}\sum\limits_{q \in \mathcal{Q}} p(y_q),
  \label{eq:vqa_score_computation}
\end{equation}
where $y_q$ is the token corresponding to the answer expected of the target image for the question $q$.

This score provides interpretable, attribute-level insights into the relevance of the candidate image with respect to the modification text, as it computes the likelihood of each requested change being correctly applied. The target image is expected to obtain a score close to 1, while an incompatible candidate image should get a lower score.

Despite checking all the characteristics mentioned in the modification text, this VQA score does not take into account the visual characteristics present only in the reference image. Thus, we combine it with the score computed by a similarity-based CIR method, which also factors visual similarity with the reference image. Hence, the CIR score gives a first ranking of the candidate images, while the VQA score lowers the rank of images incompatible with the modification text, and promotes lower-ranked images which correctly apply the requested changes.

To compute the score of a candidate image $I_i$, we take its normalized CIR score to match the range of the VQA score, and sum it with the weighted VQA score:
\begin{equation}
    x_i = \text{norm}\left(x_i^{\text{CIR}}\right) + \lambda_{VQA} \cdot \sigma_k\left( x_i^{\text{VQA}} \right),
    \label{eq:score_fusion}
\end{equation}
where $\sigma_k$ is a sigmoid-like function which makes high VQA scores more similar, in order to make the candidate image's rank more robust to eventual VQA prediction errors:
\begin{equation}
    \sigma_k : x \longmapsto \frac{1}{2} + \coth\left(\frac{1}{2k}\right) \cdot \left[ \frac{1}{1 + \exp\left(-\frac{x}{k}\right)} - \frac{1}{2} \right].
\end{equation}

Note that for efficiency purposes, we compute the VQA score only for the top-$n$ candidate images (sorted according to their CIR score), and for all $N-n$ remaining candidate images, we use their normalized CIR score.

The candidate images $I_i$ are then ranked according to their score $x_i$.
\section{Experiments}
\label{sec:experiments}

\subsection{Experimental Setup}
\label{subsec:experimental_setup}

\paragraph{Datasets.}
To construct our VQA dataset described in \Cref{subsec:vqa_finetuning}, we use the training split of the Fashion IQ~\cite{wu2020fashioniq} dataset and generate VQA questions using the process detailed in \Cref{subsec:criteria_extraction}.
We evaluate our methods using the validation split of the Fashion IQ dataset, as well as two additional derived datasets whose goal is to mitigate the annotation errors of Fashion IQ. Refined-FashionIQ~\cite{huynh2025collm} uses a LLM-based approach to keep correct Fashion IQ triplets and to re-generate the modification text of wrongly-annotated CIR triplets. enhFashionIQ~\cite{garderes2025facap} uses the images from Fashion IQ but creates new CIR triplets by randomly sampling target images based on visual similarity, and generating a modification text from the image pair, resulting in a larger CIR dataset with more precise annotations.

\paragraph{Metrics.}
We evaluate the performance of the VQA model by measuring its answer prediction accuracy. To evaluate the CIR performance, we use Recall@$k$ (with $k \in \{10;50\}$ similarly to previous works), which computes the percentage of target images that appear in the top-$k$ retrieved images list. We also use the Mean Reciprocal Rank (MRR), which gives a more granular insight in the rank of the annotated target image. All metrics are computed for each available clothing category: dress, shirt, and toptee.

\paragraph{Implementation details.}
We use GPT-5-mini\footnote{https://platform.openai.com/docs/models/gpt-5-mini} to generate the questions from the modification text. We use InternVL-3-1B~\cite{zhu2025internvl3exploringadvancedtraining} as our VQA model, and we automatically generate the annotations of our VQA dataset using InternVL-3-78B~\cite{zhu2025internvl3exploringadvancedtraining}. The VQA model is fine-tuned with a LoRA of dimension 128, on two GPUs H100, employing a batch size of 512. For the score interpolation function in \Cref{eq:score_fusion}, we set $\lambda_{VQA} = 0.068$ and $k = 0.8375$. We conduct our inference experiments on a single GPU H100. We use $n=250$ in our experiments.

\subsection{Experimental Results}
\label{subsec:experimental_results}

\begin{table}
  \caption{Statistics on the question generation process, on the Fashion IQ validation dataset. The manual evaluation is performed over 136 questions randomly sampled from the three clothing types.}
  \label{tab:info_extraction_evaluation}
  \centering
  \vspace*{-0.5\baselineskip}
  \begin{tabular}{cc|c}
    \toprule
    \makecell{Evaluation \\ type} & Statistics & Value \\
    \midrule
    \multirow{2}{*}{\makecell{Automatic}} & \makecell{Average number of \\  questions per CIR triplet} & 3.7 \\
    & Dual-image inputs & 38 \% \\
    \midrule
    \multirow{5}{*}{Manual}
    & One characteristic per question & 97 \% \\ 
    & No question repetition & 84 \% \\
    & Hallucinations & 6 \% \\
    & Correct expected answers & 100 \% \\
    & Correct number of input images & 100 \% \\
    \bottomrule
  \end{tabular}
\end{table}

\paragraph{Question generation.} Some statistics on the question generation process are given in \Cref{tab:info_extraction_evaluation}. The high amount of dual-image inputs proves that many changes described in the modification texts also need the reference image to be properly verified, hence highlighting a benefit of our approach. A manual evaluation over 136 randomly sampled questions also shows that the generated questions focus on distinct elements, and that the LLM used successfully detects the correct expected answer and whether the reference image is required to answer the question.

\begin{table}
  \caption{Evaluation of our VQA model, in its pre-trained and fine-tuned version. The automatic evaluation is performed over the Fashion IQ validation target images, while the manual evaluation is performed over a sample of 600 Fashion IQ validation images.}
  \label{tab:vqa_evaluation}
  \centering
  \vspace*{-0.5\baselineskip}
  \begin{tabular}{ccc}
    \toprule
    Model version & Automatic eval. & Manual eval. \\
    \midrule
    Pre-trained & 67.10 \% & 75.83 \% \\
    Fine-tuned & 81.92 \% & 83.67 \% \\
    \bottomrule
  \end{tabular}
\end{table}

\paragraph{Fine-tuning the VQA model.}
We evaluate the contribution of fine-tuning the VQA model in \Cref{tab:vqa_evaluation}. First, we automatically evaluate the accuracy of the VQA model before (pre-trained version) and after (fine-tuned version) fine-tuning, on the target images of the Fashion IQ validation dataset. We select these images because the question generation process from \Cref{subsec:criteria_extraction} predicts their expected answer with high fidelity, as shown in \Cref{tab:info_extraction_evaluation}. The fine-tuning process improves the accuracy by $14.82 \%$: we attain the accuracy of larger VQA models, while retaining a fast inference speed.
Second, as the target images mainly expect a “Yes” answer to the questions on the target images, we manually evaluate 600 VQA results on both target and non-target images. This confirms that the performance of the VQA model remains stable on a more balanced evaluation set. We observe a similar performance improvement between the pre-trained and fine-tuned versions of the model, hence confirming the benefits of our fine-tuning approach.

\begin{table*}
  \caption{Results on the Fashion IQ validation split for state-of-the-art CIR methods. Best results are highlighted in bold.}
  \label{tab:cir_fashioniq_evaluation}
  \centering
  
  \vspace*{-0.5\baselineskip}
  \begin{tabular}{l|cc|cc|cc|ccc}
    \toprule
    \multirow{2}{*}{\textbf{Model}} & \multicolumn{2}{c}{\textbf{Dresses}} & \multicolumn{2}{c}{\textbf{Shirts}} & \multicolumn{2}{c}{\textbf{Tops\&tees}} & \multicolumn{3}{c}{\textbf{Average}} \\
     & R@10 & R@50 & R@10 & R@50 & R@10 & R@50 & R@10 & R@50 & Global \\
    \toprule
    SQUARE~\cite{wu2025square} & 45.04 & 62.51 & 37.68 & 60.19 & 49.87 & 69.25 & 44.20 & 63.98 & 54.09 \\
    ReRanking~\cite{liu2024candidate} & 48.14 & 71.34 & 50.15 & 71.25 & 55.23 & 76.80 & 51.17 & 73.13 & 62.15 \\
    DetailFusion~\cite{yang2025detailfusion} & 51.34 & 74.05 & 58.12 & \textbf{75.95} & 61.22 & 80.09 & 56.89 & 76.70 & 66.79 \\
    \midrule
    CLIP4CIR~\cite{baldrati2022conditioned} & 39.17 & 64.20 & 44.80 & 65.36 & 47.58 & 71.09 & 43.85 & 66.88 & 55.37 \\
     \quad + VQA4CIR~\cite{feng2025vqa4cir} & 40.91 & 65.13 & 45.62 & 65.68 & 49.21 & 71.22 & 45.24 & 67.34 & 56.29 \\
     \rowcolor[HTML]{EFEFEF} 
     \quad + FIRE-CIR (ours) & 42.74 & 67.18 & 47.06 & 67.66 & 51.15 & 74.09 & 46.98 & 69.64 & 58.31 \\

     \midrule
    SPRC~\cite{bai2024sentence} & 49.03 & 72.68 & 55.05 & 74.39 & 59.15 & 78.99 & 54.41 & 75.35 & 64.88 \\
     \quad + VQA4CIR~\cite{feng2025vqa4cir} & 49.18 & 73.06 & 56.79 & 74.52 & 59.67 & 79.30 & 55.21 & 75.62 & 65.41 \\
    
    \rowcolor[HTML]{EFEFEF} 
     \quad + FIRE-CIR (ours) & 50.32 & 73.77 & 56.23 & 75.12 & 59.51 & 79.40 & 55.35 & 76.10 & 65.73 \\
     \midrule
    
    FashionBLIP-2~\cite{garderes2025facap} & 51.56 & 73.43 & 56.82 & 75.27 & 58.80 & 79.65 & 55.73 & 76.12 & 65.92 \\
     \rowcolor[HTML]{EFEFEF} 
    \quad + FIRE-CIR (ours) & \textbf{54.09} & \textbf{75.41} & \textbf{58.68} & 75.76 & \textbf{61.70} & \textbf{80.88} & \textbf{58.16} & \textbf{77.35} & \textbf{67.75} \\
    \bottomrule
  \end{tabular}
  
\end{table*}

\paragraph{Comparison to CIR state-of-the-art methods.}
We detail in \Cref{tab:cir_fashioniq_evaluation} the performance of current state-of-the-art CIR methods on the Fashion IQ dataset, with a focus on each clothing category.
We combine FIRE-CIR with the FashionBLIP-2 model, improving its average recall by 1.83\% and its average recall@10 by $2.43 \%$, surpassing the performance of DetailFusion. We note that the performance improvement is larger for the dress category compared to shirts and tops/tees. We hypothesize that this is linked to the wide diversity of dress characteristics: it leads to more specific changes mentioned in the modification text, which our method is able to leverage efficiently.
When comparing the performance of FIRE-CIR to the state-of-the-art re-ranking model VQA4CIR, we observe that FIRE-CIR obtains better recall values, whether combined with CLIP4CIR or SPRC.

\paragraph{Improving CIR methods.}
\Cref{tab:comparison_cir_reranking_methods} presents the benefits of using FIRE-CIR to improve a CIR method.
We measure the retrieval performance of three different CIR models without re-ranking (\xmark) and with our model FIRE-CIR (\cmark). We evaluate the average recall and MRR averaged over the three available clothing types (dress, shirt, toptee) on Fashion IQ, enhFashionIQ and Refined-FashionIQ.
FIRE-CIR consistently improves the performance of all three CIR models tested here, across the three datasets and on both metrics. We note that the performance improvement is more significant on enhFashionIQ and Refined-FashionIQ, probably due to the more detailed and specific modification texts, which help FIRE-CIR focus on the relevant elements to detect in the candidate images.

\begin{table*}
  \caption{
  CIR performance improvement when combining a CIR method with our FIRE-CIR model. For each initial CIR model, the best results are highlighted in bold.
  }
  \label{tab:comparison_cir_reranking_methods}
  \centering
  \vspace*{-0.5\baselineskip}

  \begin{tabular}{c|c|cc|cc|cc}
    \toprule
    \multirow{2}{*}{\textbf{Model}} & \multirow{2}{*}{\textbf{\makecell{FIRE-CIR \\ activated}}} & \multicolumn{2}{c}{\textbf{Fashion IQ}} & \multicolumn{2}{c}{\textbf{enhFashionIQ}} & \multicolumn{2}{c}{\textbf{Refined-FashionIQ}} \\
      & & Recall & MRR & Recall & MRR & Recall & MRR \\
    \toprule
    \multirow{2}{*}{CLIP4CIR~\cite{baldrati2022conditioned}} & \xmark & 55.37 & 0.2403 & 66.85 & 0.2737 & 73.89 & 0.4028 \\
    & \cmark & \textbf{58.31} & \textbf{0.2746} & \textbf{73.87} & \textbf{0.3604} & \textbf{78.09} & \textbf{0.4651} \\
    \midrule
    \multirow{2}{*}{SPRC~\cite{bai2024sentence}} & \xmark & 64.88 & 0.3343 & 78.33 & 0.4515 & 82.20 & 0.5246 \\
    & \cmark & \textbf{65.73} & \textbf{0.3454} & \textbf{80.02} & \textbf{0.4779} & \textbf{83.24} & \textbf{0.5459} \\
    \midrule
    \multirow{2}{*}{FashionBLIP-2~\cite{garderes2025facap}} & \xmark & 65.92 & 0.3325 & 87.10 & 0.5251 & 82.02 & 0.5050 \\
    & \cmark & \textbf{67.75} & \textbf{0.3658} & \textbf{87.99} & \textbf{0.5822} & \textbf{84.01} & \textbf{0.5617} \\
    \bottomrule
    \end{tabular}
\end{table*}

\subsection{Ablation Study \& Analysis}
\label{subsec:ablation_study_analysis}

\begin{figure}[t]
  \centering
   \includegraphics[width=\linewidth]{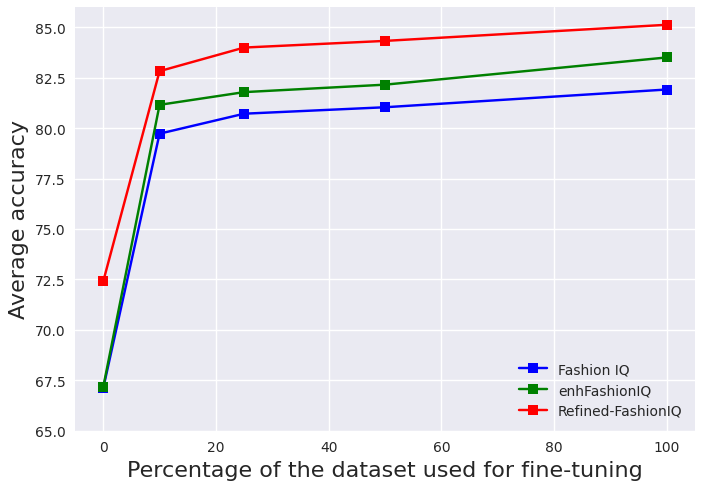}

   \caption{Accuracy of the fine-tuned VQA model depending on the percentage of our automatically-annotated dataset used. For each model version, we measure the accuracy over all the target images in our three different validation sets.}

   \label{fig:vqa_dataset_ablation}
\end{figure}

\paragraph{Dataset contribution.}
We investigate the impact of data quantity by fine-tuning the VQA model on progressively larger subsets of our automatically-annotated dataset, and evaluate the resulting models on Fashion IQ. As shown in \Cref{fig:vqa_dataset_ablation}, the performance of the model increases considerably when fine-tuning the model even on a smaller-scale dataset, but performance gains with larger subsets are smaller. Further improvements may require focusing on fashion concept diversity rather than dataset volume.

\begin{table*}
  \caption{Ablation study of the FIRE-CIR model. 
  The best and second-best results are highlighted in bold and underlined, respectively.
  }
  \label{tab:ablation_study_pipeline}
  \centering
  \vspace*{-0.5\baselineskip}
  \begin{tabular}{ccc|cccc}
    \toprule
    \multirow{2}{*}{\textbf{\makecell{Multiple \\ questions}}} & \multirow{2}{*}{\textbf{\makecell{VQA model \\ fine-tuning}}} & \multirow{2}{*}{\textbf{\makecell{With score \\ merging method}}} & \multicolumn{4}{c}{\textbf{Fashion IQ}} \\
     & & & Recall@10 & Recall@50 & Avg. recall & MRR \\
    \toprule
    \xmark & \xmark & \xmark & 56.44 & 76.78 & 66.61 & 0.3513 \\
    \cmark & \xmark & \xmark & 57.28 & 77.01 & 67.14 & 0.3520 \\
    \cmark & \cmark & \xmark & \underline{57.43} & \underline{77.30} & \underline{67.36} & \underline{0.3584} \\
    \cmark & \cmark & \cmark & \textbf{58.16} & \textbf{77.35} & \textbf{67.75} & \textbf{0.3658} \\
    \bottomrule
  \end{tabular}
  
\end{table*}

\paragraph{Component ablation.}
To highlight the benefits of our contributions, we evaluate our method in four settings in \Cref{tab:ablation_study_pipeline}. The first row corresponds to the baseline, prompting the VLM to globally evaluate the relevance of a candidate image with respect to the modification text. 
The second row shows the performance when decomposing the relevance analysis in multiple questions.
The third row illustrates the benefits of fine-tuning the VQA model on our automatically-annotated dataset, for domain adaptation.
Finally, the last row integrates our score merging method with $\sigma_k$. 
Each component helps the model improve its CIR performance, as displayed by the recall and MRR values when evaluating the models on the Fashion IQ dataset. The increase is more significant for the recall@10 value than the recall@50 one, highlighting that FIRE-CIR's components help the model refine its top-retrieved results.

\begin{table}[h]

  \caption{Evaluation of VQA models on our manually annotated dataset. `FT' denotes fine-tuning on the proposed training data.
  }
  \vspace*{-0.5\baselineskip}
  \label{tab:rebuttal_vqa_evaluation}
  \tabcolsep=0.08cm
  \centering
  \begin{tabular}{lcccc}
    \toprule
    Model & Precision & Recall & AUC-PR & AUC-ROC \\
    \midrule
    LLaVA-13B & 75.7 & 88.8 & 84.4 & 78.1 \\
    InternVL3-1B & 80.0 & 84.2 & 85.3 & 79.7 \\
    InternVL3-1B+FT & \textbf{84.1} & \textbf{92.7} & \textbf{93.3} & \textbf{91.0} \\
    \bottomrule
  \end{tabular}
\end{table}

\paragraph{Impact of the VLM backbone.}
We further analyze the impact of different VLM backbones to ensure a fair comparison with VQA4CIR~\cite{feng2025vqa4cir}. Specifically, VQA4CIR adopts LLaVA-1.5-13B~\cite{liu2024improved}, whereas our method leverages the more recent InternVL-3-1B~\cite{zhu2025internvl3exploringadvancedtraining} model. As shown in Table \ref{tab:rebuttal_vqa_evaluation}, without pre-training on our proposed VQA dataset, LLaVA-1.5-13B and InternVL-3-1B achieve comparable performance. This suggests that our improvements over VQA4CIR are not attributable to a stronger VLM backbone. Moreover, the 1B model enables about three times faster inference. Fine-tuning it on our VQA dataset leads to a substantial performance gain on the VQA task.

\begin{figure}[t]
  \centering
   \includegraphics[width=\linewidth]{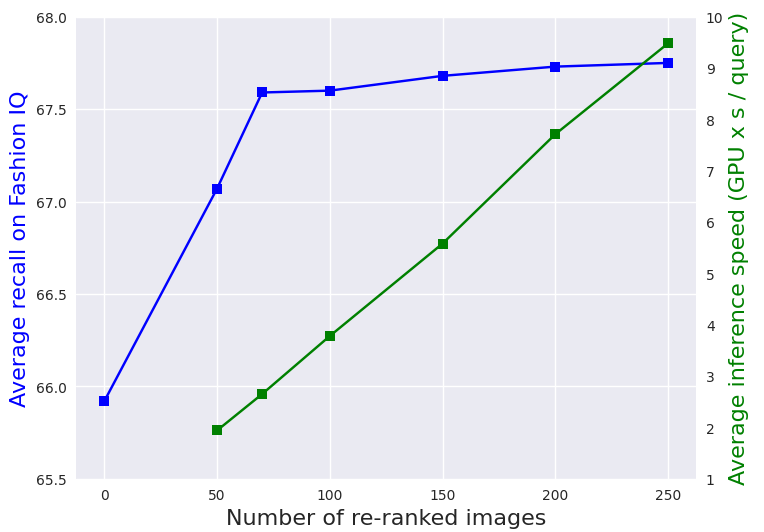}

   \caption{Average recall and inference speed of FIRE-CIR depending on the number $n$ of re-ranked images, evaluated on the Fashion IQ validation set.}
   \label{fig:ablation_study_number_images}
\end{figure}

\paragraph{Speed-performance tradeoff.}
Finally, we measure how the number $n$ of candidate images whose VQA score is computed impacts both CIR performance and inference speed.
As shown in \Cref{fig:ablation_study_number_images}, re-ranking a larger number $n$ 
of candidate images 
leads to a better performance, at the cost of a slower inference. In our experiments, the best CIR performance is reached with $n=250$, taking about 9.5 seconds per query on one GPU. Note that the inference speed scales linearly with the number of GPUs used. 
If both inference speed and the number of GPUs is an issue, the number of re-ranked images can be reduced to $n=70$ to lower the computational requirements for practical usage (2.65 GPU second per query), while significantly improving the performance of the traditional CIR method.

\subsection{Qualitative Results}
\label{subsec:qualitative_results}

\begin{figure}[t]
  \centering

   \includegraphics[width=\linewidth]{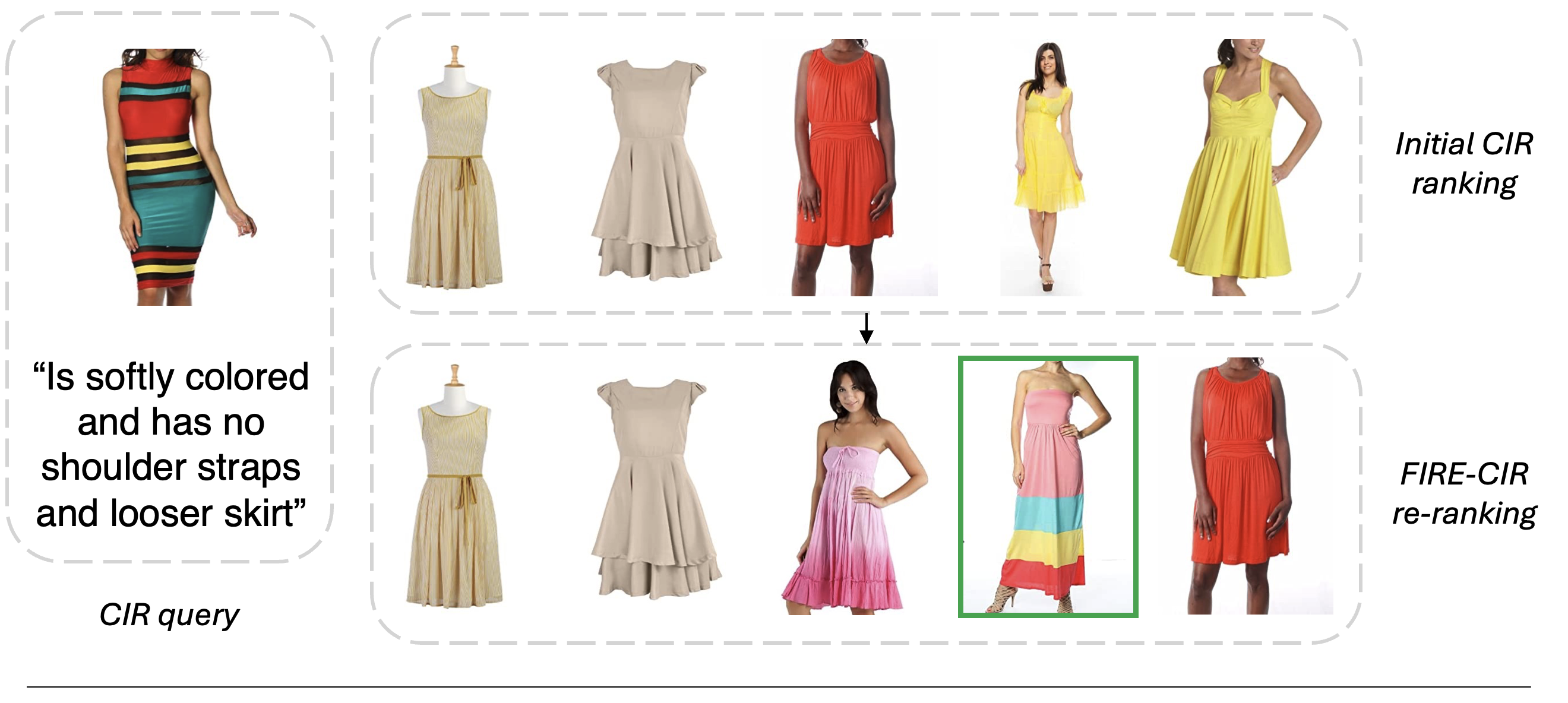}
   \includegraphics[width=\linewidth]{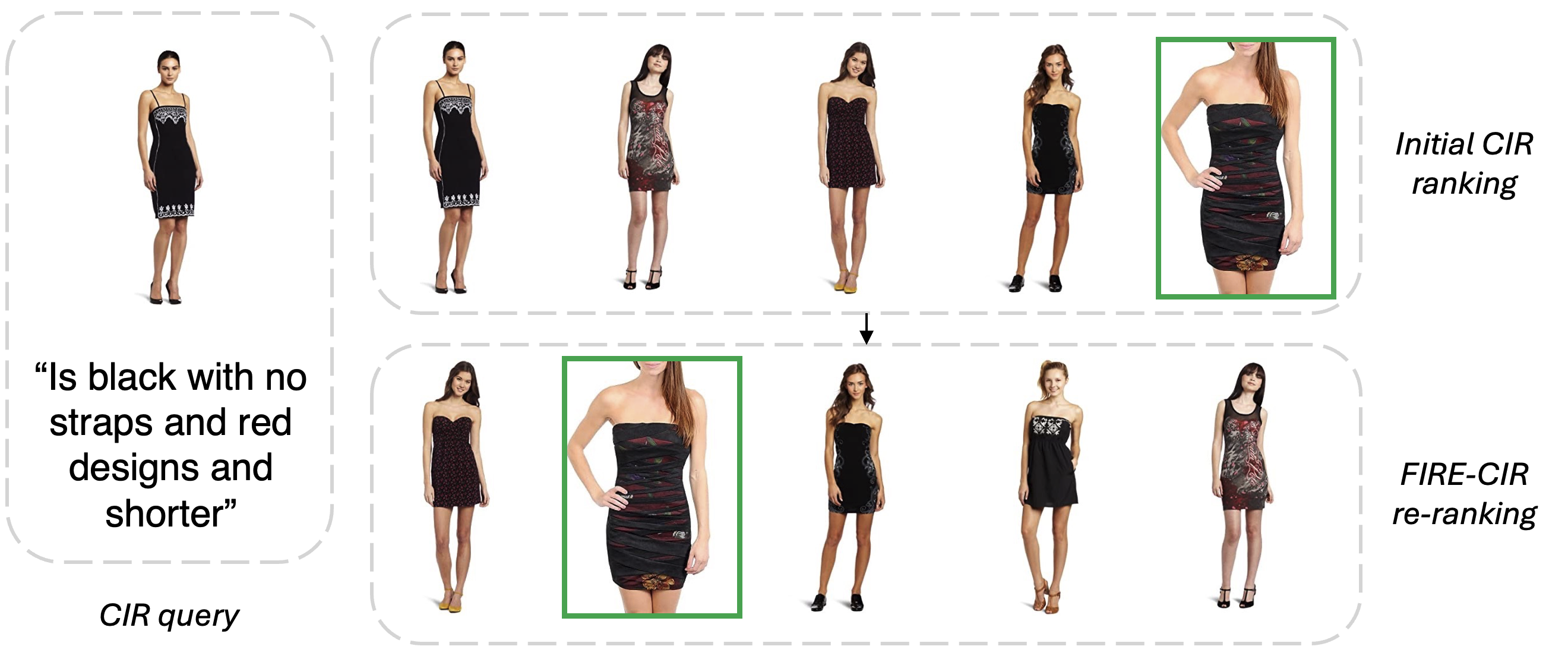}

   \caption{Qualitative examples of re-ranking with FIRE-CIR on the Fashion IQ dataset. We show the top-5 images (left to right) retrieved by each method (FashionBLIP-2 and FIRE-CIR). The ground-truth target image is framed in green.}
   \label{fig:cir_qualitative_examples}
\end{figure}

We present two qualitative examples of FIRE-CIR re-ranking on Fashion IQ in \Cref{fig:cir_qualitative_examples}. 
In the first example, the generated questions are: ``Is the dress softly colored?", ``Does the dress have no shoulder straps?" and ``Is the skirt looser than in the reference image?". As the three last images in the top-5 are not softly colored and have shoulder straps, their VQA score lowers their rank, hence promoting two other images fitting better the modification text, including the target image. 
Similarly, in the second example, the generated questions are: ``Is the dress black?", ``Is the dress strapless?", ``Does the dress have red designs?", ``Is the dress shorter than the reference image?". The retrieved images with straps are pushed lower in the ranking, while the target and other relevant images are given a higher rank. In both cases, the VQA score of the target image is 1.

\section{Conclusion}
\label{sec:conclusion}

In this paper, we propose a new model bringing compositional reasoning and interpretability into fashion CIR. FIRE-CIR first decomposes the modification text into a set of visual questions, to check the relevance of a given candidate image. These questions are then answered using a VQA model, fine-tuned on an automatically-annotated dataset to adapt it to the fashion domain. This reasoning system computes a score, combinable with another CIR method to improve its performance.
Qualitative and quantitative experiments show that FIRE-CIR consistently improves the performance of CIR models, and achieves state-of-the-art performance on the Fashion IQ benchmark. Moreover, this model can adapt to different applications, as its performance can be lowered slightly to improve the inference speed.
For future work, the reasoning system could benefit from including visual attributes not impacted by the changes, so that additional constraints can be checked in the candidate images to better filter irrelevant ones.

\section*{Acknowledgement}
This project was granted access to the HPC resources of IDRIS under the allocation AD011015247R1 made by GENCI. It was funded in part by the French government under management of Agence Nationale de la Recherche as part of the "France 2030" program, reference ANR-23-IACL-0008 (PR[AI]RIE-PSAI project), and Paris Île-de-France Région in the frame of the DIM AI4IDF.
    
{
    \small
    \bibliographystyle{ieeenat_fullname}
    \bibliography{main}
}

\appendix
\clearpage

\section{Additional FIRE-CIR qualitative examples}
\label{sec:supplementary_qualitative_examples}

\begin{figure*}[t]
  \centering
  
   \includegraphics[width=.95\linewidth]{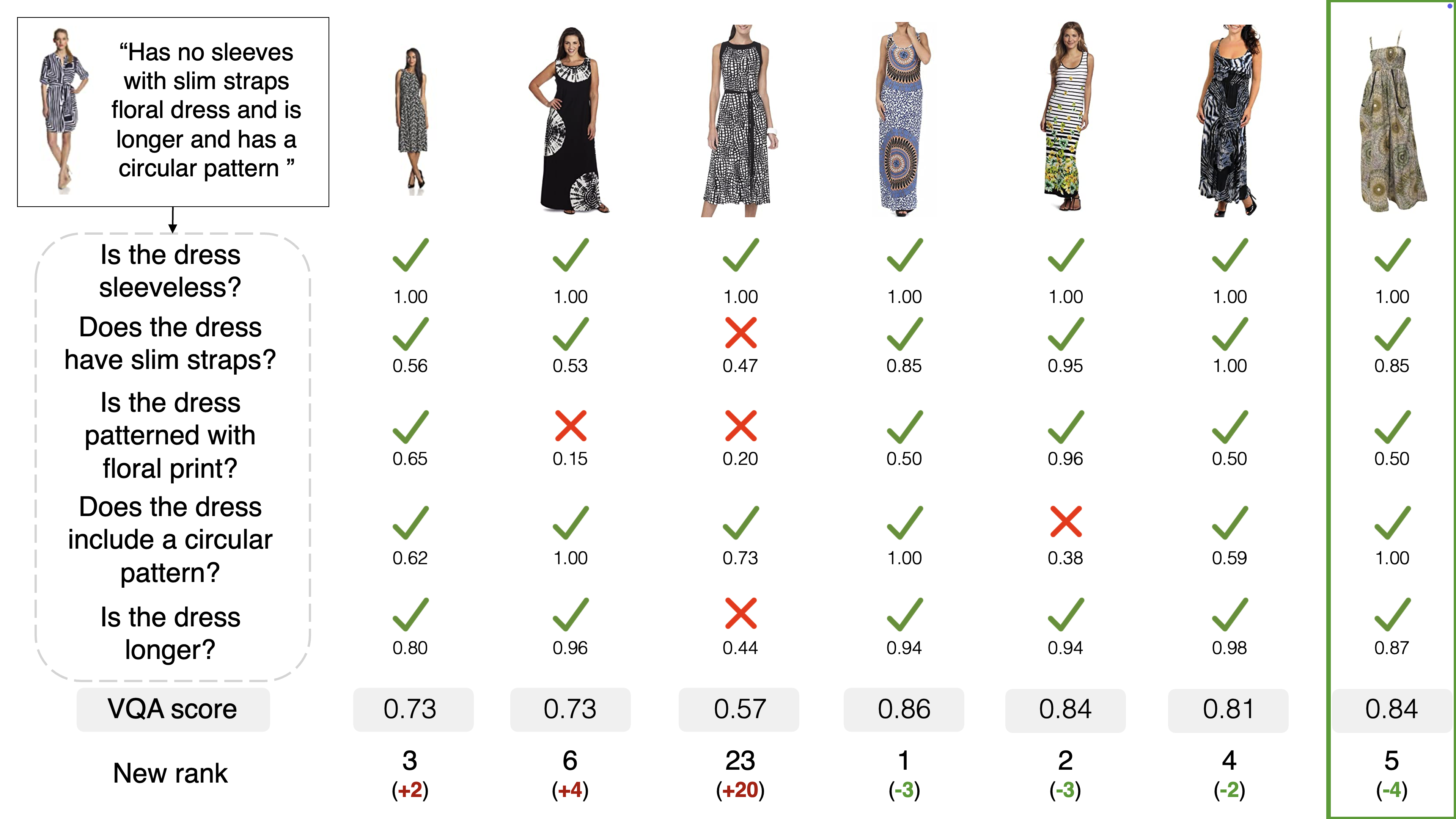}
  
   \caption{In this example, FIRE-CIR is able to accurately identify the pattern in the dresses and re-ranks the retrieved results according to their resemblance with what is described in the modification text.}
   \label{fig:supplementary_qualitative_examples_1}
\end{figure*}

\begin{figure*}[t]
  \centering

   \includegraphics[width=.95\linewidth]{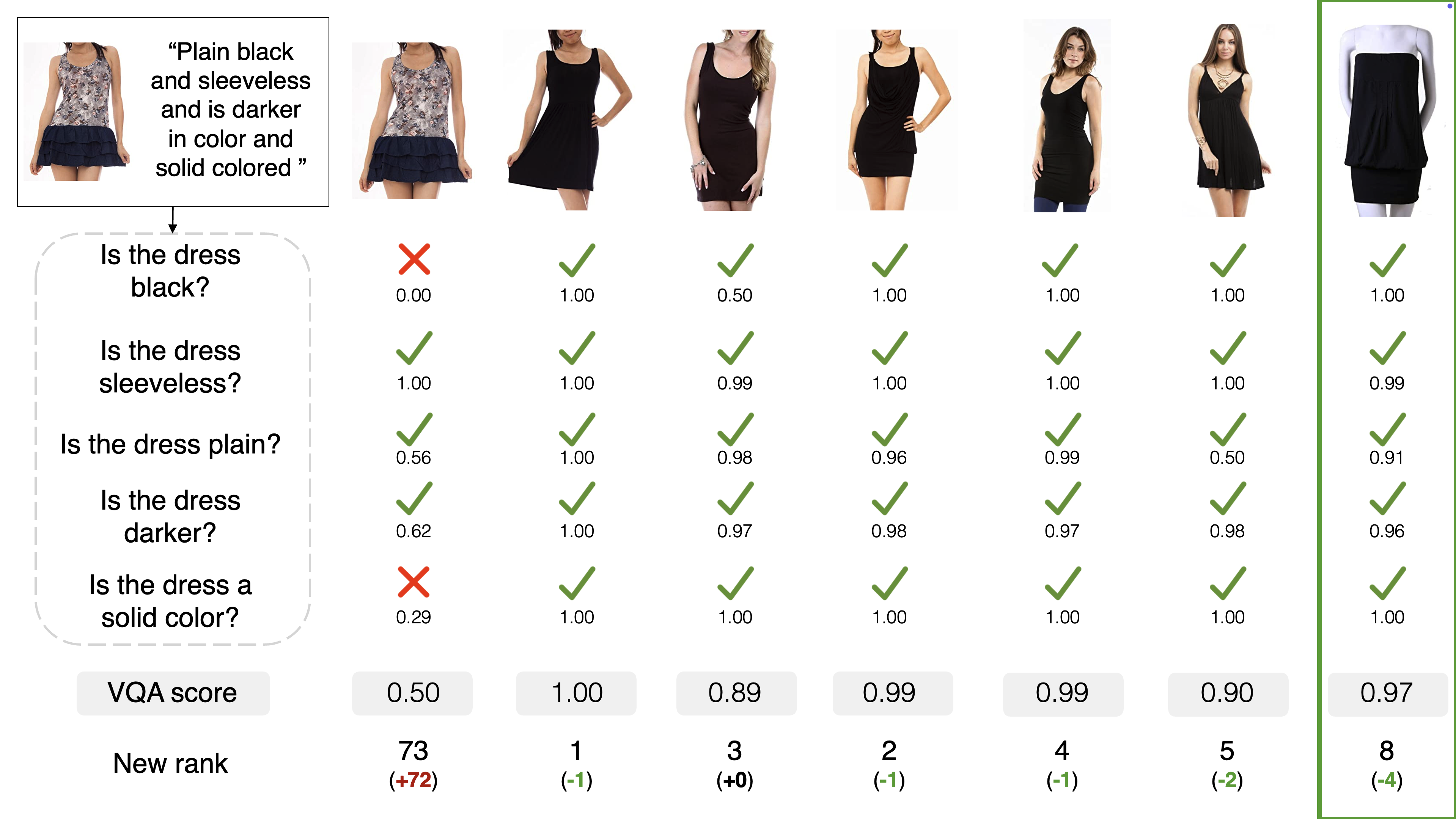}

   \caption{While FashionBLIP-2 ranks highly the reference image as some of its visual aspects remain compatible with the CIR query, FIRE-CIR correctly identifies that this dress does not have a solid black color, and thus re-ranks it outside of the top-50 results. The ground-truth target image is at the 8th rank, but the 7 other top-ranked images are also compatible with the given query.}
   \label{fig:supplementary_qualitative_examples_2}
\end{figure*}

\begin{figure*}[t]
  \centering
  
   \includegraphics[width=.95\linewidth]{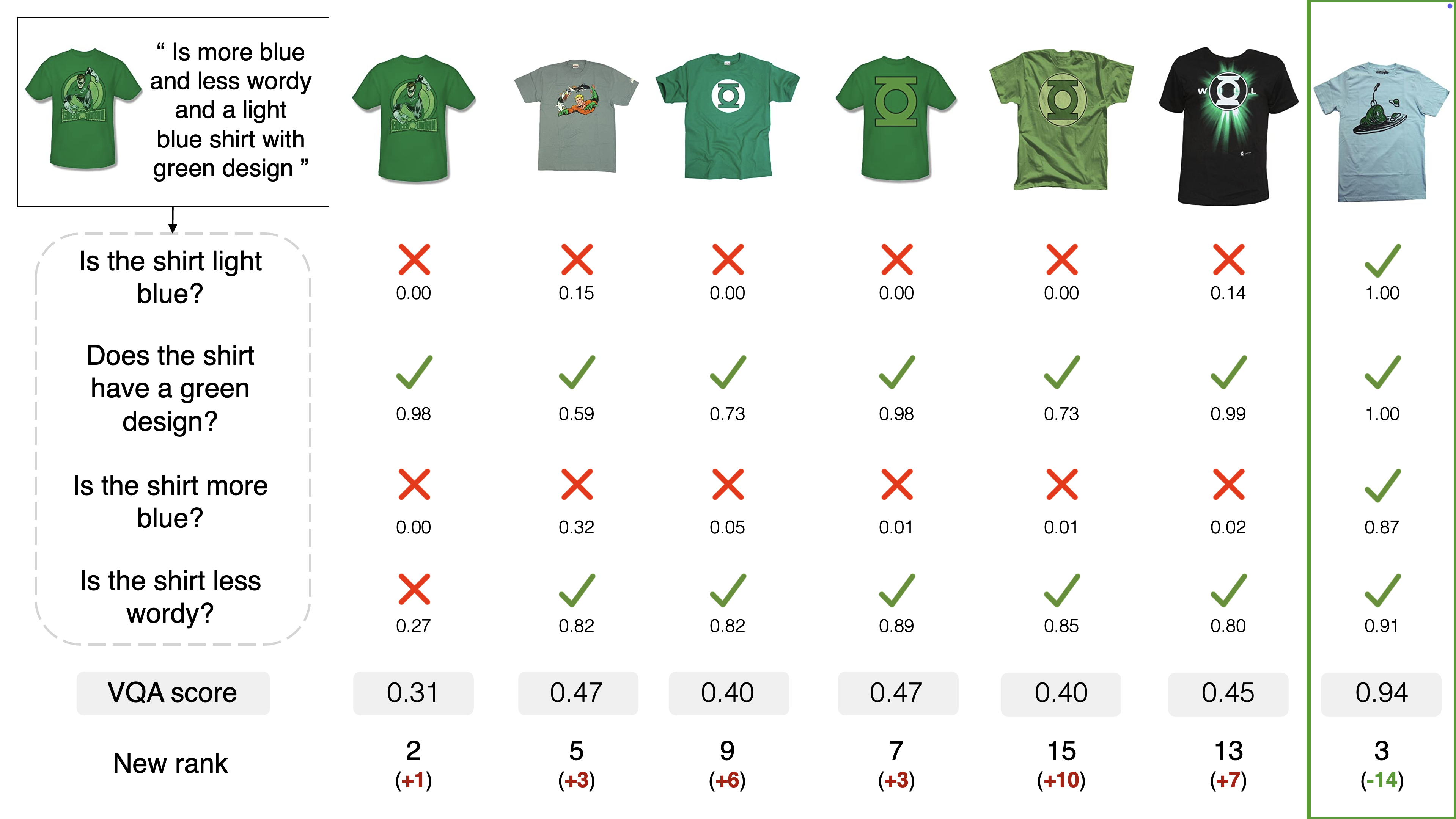}
  
   \caption{Similarly, visual similarity and the green design contribute to having candidate images similar to the reference one in the top-retrieved results. However, FIRE-CIR accurately promotes the ground-truth image, as its color is more coherent with what is written in the modification text.}
   \label{fig:supplementary_qualitative_examples_3}
\end{figure*}

\begin{figure*}[t]
  \centering
  
   \includegraphics[width=.95\linewidth]{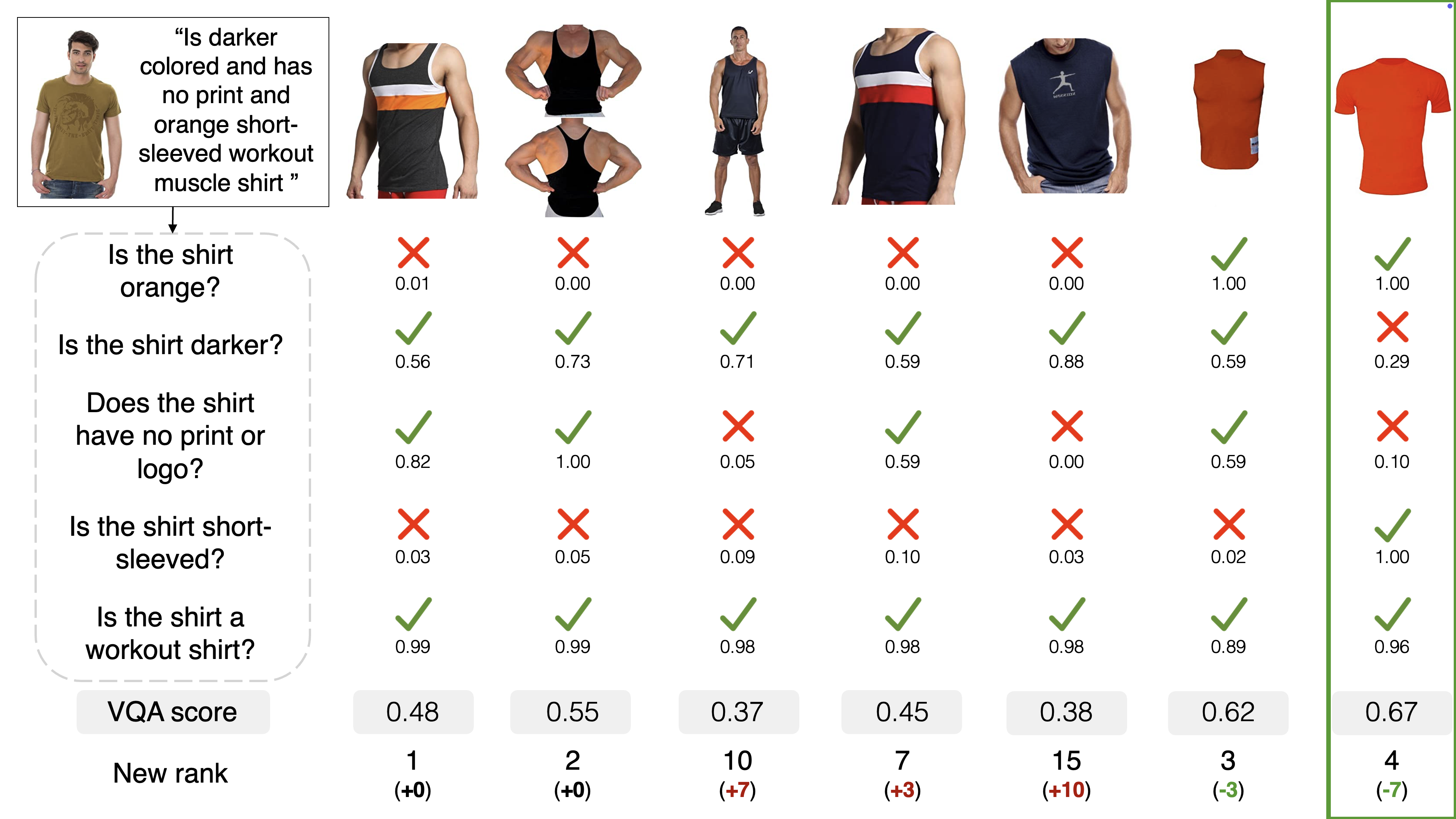}

   \caption{Contrary to FashionBLIP-2 which focuses specifically on the ``workout muscle" characteristic, FIRE-CIR gives more importance to all the features, including the orange color. }
   \label{fig:supplementary_qualitative_examples_4}
\end{figure*}

To further illustrate the reasoning process, we present additional qualitative examples of FIRE-CIR on the Fashion IQ dataset (``dress" subset for \Cref{fig:supplementary_qualitative_examples_1} and \Cref{fig:supplementary_qualitative_examples_2}, and ``shirt" subset for \Cref{fig:supplementary_qualitative_examples_3} and \Cref{fig:supplementary_qualitative_examples_4}). 

For each example, the reference image and modification text are in the top-left corner. FIRE-CIR decomposes the modification text into a set of visual questions, listed on the left. These questions are then applied to the top-6 images retrieved by FashionBLIP-2 (from the first image on the top left to the sixth on the top right), and to the ground-truth target image (top-right corner). Each checkmark (resp. cross) corresponds to a predicted answer compatible (resp. incompatible) with the modification text. The probability of the answer being compatible with the text is displayed below the checkmark (resp. cross). Then, all answer probabilities are averaged for each candidate image to compute the VQA score, which is used to compute the final rank of each candidate image. The final rank is indicated at the bottom of each example. We also specify the rank difference with the FashionBLIP-2 results: a negative difference means that the candidate image gets a higher rank (so it is more relevant), while a positive difference means that the candidate image is further among the retrieved results (so it is less relevant).


\end{document}